\pdfoutput=1

\documentclass[11pt,a4paper]{article}
\PassOptionsToPackage{hyphens}{url}\usepackage[hyperref]{naaclhlt2018}
\usepackage{latexsym}
\usepackage{times}
\usepackage{url}
\usepackage{latexsym}
\usepackage{mathtools}
\usepackage{amsthm}
\usepackage{color}
\usepackage{amssymb}
\usepackage{pifont}
\usepackage{epstopdf}
\usepackage{enumitem}

\newcommand{\qt}[1]{{\textquoteleft {#1}\textquoteright}}

\renewcommand{\vec}[1]{\mathbf{#1}}

\aclfinalcopy 

\title{A Transition-based Algorithm for Unrestricted 
AMR Parsing}

\author{David Vilares \\
  Universidade da Coru\~{n}a \\
  FASTPARSE Lab, LyS Group \\
  Departamento de Computaci\'{o}n \\
  Campus de A Elvi\~{n}a s/n, 15071 \\ A Coru\~{n}a, Spain \\
  {\tt david.vilares@udc.es} \\
  \\\And
  Carlos G\'{o}mez-Rodr\'{i}guez \\
  Universidade da Coru\~{n}a \\
  FASTPARSE Lab, LyS Group \\
  Departamento de Computaci\'{o}n \\
  Campus de A Elvi\~{n}a s/n, 15071 \\ A Coru\~{n}a, Spain \\
  {\tt carlos.gomez@udc.es} \\}

\date{}

\begin{document}
\maketitle
\begin{abstract}
Non-projective parsing can be useful to handle cycles and reentrancy in \textsc{amr} graphs. We explore this idea and introduce a greedy left-to-right non-projective transition-based parser. At each parsing configuration, an oracle decides whether to create a concept  or whether to connect a pair of existing concepts. 
The algorithm handles reentrancy and arbitrary cycles natively, i.e. within the transition system itself.
The model is evaluated on the LDC2015E86 corpus, obtaining results close to the state of the art, including a Smatch of 64\%, and showing good behavior on reentrant edges. 
\end{abstract}



\section{Introduction}

\begin{figure}[hbtp]
\centering
\includegraphics[width=1\columnwidth]{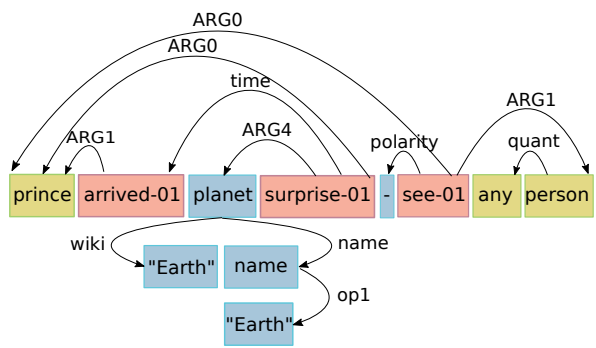}
\caption{\label{f-amr-example} \textsc{amr} graph for \qt{When the prince arrived on the Earth, he was surprised not to see any people}. Words can refer to concepts by themselves (green), be mapped to PropBank framesets (red) or be broken down into multiple-term/non-literal concepts (blue). \texttt{Prince} plays different semantic roles.}
\end{figure}

Abstract Meaning Representation (\textsc{amr}) is a semantic representation language to map the meaning of English sentences into directed, cycled, labeled graphs \cite{banarescu2013abstract}. 
Graph vertices are concepts
inferred from words. The concepts can be represented by the words themselves (e.g. \texttt{dog}), PropBank framesets \cite{palmer2005proposition} (e.g. \texttt{eat-01}), or keywords 
(like named entities or quantities).
The edges denote 
relations 
between pairs of concepts (e.g. \texttt{eat-01 :ARG0 dog}).
\textsc{amr} parsing integrates tasks that have usually been addressed separately in natural language processing (\textsc{nlp}), such as named entity recognition \cite{nadeau2007survey}, semantic role labeling \cite{palmer2010semantic} or co-reference resolution \cite{ng2002improving,lee2017scaffolding}. Figure \ref{f-amr-example} shows an example of an \textsc{amr} graph.

Several transition-based dependency parsing algorithms have been extended to generate \textsc{amr}. \newcite{wang2015transition} describe a two-stage model, where they first obtain the dependency parse of a sentence and then transform it into a graph. \newcite{damonte-cohen-satta:2017:EACLlong} propose a variant of the \textsc{arc-eager} algorithm to identify  labeled edges between concepts. These concepts are identified using a lookup table and a set of rules. 
A restricted subset of reentrant edges are supported by an additional classifier. 
A similar configuration is used in \cite{gildea-satta-cl17,peng-aaai18}, but relying on a cache data structure to handle reentrancy, cycles and restricted non-projectivity. A feed-forward network and additional hooks are used to build the concepts. \newcite{ballesteros-alonaizan:2017:EMNLP2017} use a modified \textsc{arc-standard} algorithm, where the oracle is trained using stack-\textsc{lstm}s \cite{dyer-EtAl:2015:ACL-IJCNLP}.
Reentrancy is handled through \textsc{swap} \cite{nivre2009non} and they define additional transitions intended to detect concepts, entities and polarity nodes.

This paper explores unrestricted non-projective \textsc{amr} parsing and introduces \textsc{amr-covington}, inspired by \newcite{covington2001fundamental}. It handles arbitrary non-projectivity, cycles and reentrancy in a natural way, as there is no need for specific transitions, but just the removal of restrictions from the original algorithm. The algorithm has full coverage and keeps transitions simple, which is a matter of concern in recent studies 
\cite{peng-aaai18}.

\section{Preliminaries and Notation}

\paragraph{Notation} We use \texttt{typewriter} font for concepts and their indexes (e.g. \texttt{dog} or \texttt{1}), regular font for raw words (e.g. dog or 1), and a bold style font for vectors and matrices (e.g. $\vec{v}$, $\vec{W}$).

\noindent\newcite{covington2001fundamental} describes a fundamental algorithm for unrestricted  non-projective dependency parsing. The algorithm can be implemented as a left-to-right transition system \cite{nivre2008algorithms}. The key idea is intuitive. Given a word to be processed at a particular state, the word is compared against the words that have previously been processed, deciding to establish or not a syntactic dependency arc from/to each of them. The process continues until 
all previous words are checked
or until the algorithm decides no more connections with  previous words need to be built, then the next word is processed. The runtime is $\mathcal{O}(n^2)$ in the worst scenario. 
To guarantee the single-head and acyclicity conditions that are required in dependency parsing, explicit tests are added to the algorithm to check for transitions that would break the constraints. These are then disallowed, making the implementation less straightforward.

\section{The \textsc{amr}-Covington algorithm}

The acyclicity and single-head constraints are not needed in \textsc{amr}, as arbitrary graphs are allowed.
Cycles and reentrancy are used to model semantic relations between concepts (as shown in  Figure \ref{f-amr-example}) and to identify co-references. By removing the constraints from the Covington transition system, we achieve a natural way to deal with them.\footnote{This is roughly equivalent to going back to the  naive parser called ESH in \cite{covington2001fundamental}, which has not seen practical use in parsing due to the lack of these constraints.} 

Also, \textsc{amr} parsing requires words to be transformed into concepts. Dependency parsing operates on a constant-length sequence. But in \textsc{amr}, 
words can be removed, generate a single concept, or generate several concepts.
In this paper, additional lookup tables and transitions are defined to 
create concepts when needed,
following the current trend \cite{damonte-cohen-satta:2017:EACLlong,ballesteros-alonaizan:2017:EMNLP2017,gildea-satta-cl17}.

\subsection{Formalization}\label{section-formalization}

Let $G$=$(V,E)$ be an edge-labeled directed graph where: $V$=$\{\texttt{0},\texttt{1},\texttt{2},\ldots,\texttt{M}\}$ is the set of concepts and $E = V \times E \times V$ is the set of labeled edges, we will 
denote a connection between a head concept $\texttt{i} \in V$ and a dependent concept $\texttt{j} \in V$ as $\texttt{i} \xrightarrow{l} \texttt{j}$, where \emph{l} is the semantic label connecting them.

The parser will process sentences from left to right. Each 
decision leads to a new parsing configuration, which 
can be abstracted as a 4-tuple $(\lambda_1,\lambda_2,\beta,E)$ where:
\begin{itemize}
\item $\beta$ is a buffer that contains unprocessed words. They await to be transformed to a concept, a part of a larger concept, or to be removed. In $b|\beta$, $b$ represents the head of $\beta$, and it optionally can be a concept. In that case, it will be denoted as \texttt{b}. 
\item $\lambda_1$ is a list of previously created concepts that are waiting to determine its semantic relation with respect to \texttt{b}. Elements in $\lambda_1$ are concepts. In $\lambda_1|\texttt{i}$, $\texttt{i}$ denotes its last element.
\item $\lambda_2$ is a list that contains previously created concepts for which the relation with \texttt{b} has already been determined. Elements in $\lambda_2$ are concepts.  In $\texttt{j}|\lambda_2$, $\texttt{j}$ denotes the head of $\lambda_2$.
\item $E$ is the set of the created edges.
\end{itemize}

\begin{table*}[h]
\begin{center}
\begin{tabular}{lll}
 \bf Transitions & \bf Step $t$ & \bf Step $t+1$ \\ 

{\sc left-arc$_l$} & $(\lambda_1|\texttt{i}, \lambda_2, \texttt{b}|\beta, E)$ & $(\lambda_1, \texttt{i}| \lambda_2, \texttt{b}|\beta, E \cup \{\texttt{b} \xrightarrow{l} \texttt{i}\})$  \\

{\sc right-arc$_l$} & $(\lambda_1|\texttt{i}, \lambda_2,\texttt{b}|\beta,E)$ & $(\lambda_1,\texttt{i} |\lambda_2,\texttt{b}|\beta,E \cup \{\texttt{i} \xrightarrow{l} \texttt{b}\})$  \\

{\sc multiple-arc$_{l_1,l_2}$}&$(\lambda_1|\texttt{i}, \lambda_2, \texttt{b}|\beta, E)$ & $(\lambda_1, \texttt{i}| \lambda_2, \texttt{b}|\beta, E \cup \{\texttt{b} \xrightarrow{l} \texttt{i}\} \cup \{\texttt{i} \xrightarrow{l_2} \texttt{b}\})$  \\

{\sc shift} & $(\lambda_1,\lambda_2,\texttt{b}|\beta,E)$ & $(\lambda_1 \cdot \lambda_2|\texttt{b},[],\beta,E)$  \\

{\sc no-arc} & $(\lambda_1|\texttt{i}, \lambda_2,\beta,E)$ & $(\lambda_1, \texttt{i}|\lambda_2, \beta, E)$ \\

{\sc confirm} & $(\lambda_1, \lambda_2, b|\beta,E)$ & $(\lambda_1, \lambda_2, \texttt{b}|\beta, E)$ \\
{\sc breakdown}$_\alpha$ &$(\lambda_1, \lambda_2,b|\beta,E)$&$(\lambda_1, \lambda_2,\texttt{b}|b|\beta, E)$\\ 
{\sc reduce}&$(\lambda_1, \lambda_2,b|\beta,E)$&$(\lambda_1, \lambda_2,\beta,E)$\\


\end{tabular}
\end{center}
\caption{\label{covington-transitions} Transitions for \textsc{amr-covington}}
\end{table*}

Given an input sentence, the parser starts at an initial configuration $c_s$ = $([\texttt{0}],[],1|\beta, \{\})$ and will apply valid transitions until a final configuration $c_f$ is reached, such that $c_f$ = $(\lambda_1, \lambda_2, [], E)$. The set of transitions is formally defined in Table \ref{covington-transitions}:

\begin{itemize}
\item \textsc{left-arc}$_l$: Creates an edge $\texttt{b} \xrightarrow{l} \texttt{i}$. \texttt{i} is moved to $\lambda_2$.
\item \textsc{right-arc}$_l$: Creates an edge $\texttt{i}  \xrightarrow{l} \texttt{b}$. $\texttt{i}$ is moved to $\lambda_2$.
\item \textsc{shift}: Pops \texttt{b} from $\beta$. $\lambda_1$, $\lambda_2$ and \texttt{b} are appended.
\item \textsc{no arc}: It is applied when the algorithm determines that there is no semantic relationship between \texttt{i} and \texttt{b}, but there is a relationship between some other node in $\lambda_1$ and \texttt{b}.
\item \textsc{confirm}: Pops $b$ from $\beta$ and puts the concept \texttt{b} in its place. This transition is called to handle words that only need to generate one (more) concept.
\item \textsc{breakdown}$_\alpha$: Creates a concept \texttt{b} from $b$, and places it on top of $\beta$, but $b$ is not popped, and the new buffer state is $\texttt{b}|b|\beta$. It is used to handle a word that is going to be mapped to multiple concepts. 
To guarantee termination, \textsc{breakdown} is parametrized with a constant $\alpha$, banning generation of more than $\alpha$ consecutive concepts by using this operation. Otherwise, concepts could be generated indefinitely without emptying $\beta$.

\item \textsc{reduce}: Pops $b$ from $\beta$. It is used to remove words that do not add any meaning to the sentence and are not part of the \textsc{amr} graph.
\end{itemize}

\textsc{left} and \textsc{right-arc} handle cycles and reentrancy with the exception of cycles of length 2 (which only involve \texttt{i} and \texttt{b}). To assure full coverage, we include an additional transition: \textsc{multiple-arc}$_{(l_{1},l_{2})}$ that creates two edges $\texttt{b} \xrightarrow{l_1} \texttt{i}$ and $\texttt{i}  \xrightarrow{l_2} \texttt{b}$. \texttt{i} is moved to $\lambda_2$. \textsc{multiple-arc}s are marginal and will not be learned in practice. \textsc{amr-covington} can be implemented without \textsc{multiple-arc}, by keeping \texttt{i} in $\lambda_1$ after creating an arc and using \textsc{no-arc} when the parser has finished  creating connections between \texttt{i} and \texttt{b}, at a cost to efficiency as transition sequences would be longer. 
Multiple edges in the same direction between \texttt{i} and \texttt{b} are handled by representing them as a single edge that merges the labels.

\paragraph{Example} Table \ref{table-gold-transitions-examples} illustrates a valid transition sequence to obtain the \textsc{amr} graph of Figure \ref{f-amr-example}. 

\begin{table}[t!]
\centering
\tabcolsep=0.15cm
\small{
\begin{tabular}{rrrr}
$\mathbf{\lambda_1}$ & $\mathbf{\lambda_2}$  &$\mathbf{\beta}$&\bf Action (times)\\ 
&&w, t, p&\textsc{reduce}$\times 2_1$\\
&&p, a, o&\textsc{confirm}$_2$\\
&&\texttt{p}, a, o&\textsc{shift}$_3$\\
\texttt{p}&&a, o, t&\textsc{confirm}$_4$\\
\texttt{p}&&\texttt{a}, o, t&\textsc{left-arc}$_5$\\
&\texttt{p}&\texttt{a}, o, t&\textsc{shift}$_6$\\
\texttt{p}, \texttt{a}&&o, t, E&\textsc{reduce}$\times 2_7$\\
\texttt{p}, \texttt{a}&&E, h, w&\textsc{breakdown}$_8$\\
\texttt{p}, \texttt{a}&&\texttt{\qt{E}}, E, h&\textsc{shift}$_9$\\
\texttt{p}, \texttt{a}, \texttt{\qt{E}}&&E, h, w&\textsc{breakdown}$_{10}$\\
\texttt{p}, \texttt{a}, \texttt{\qt{E}}&&\texttt{\qt{E}}, h, w&\textsc{shift}$_{11}$\\
\texttt{a}, \texttt{\qt{E}}, \texttt{\qt{E}}&&E, h, w&\textsc{breakdown}$_{12}$\\
\texttt{a}, \texttt{\qt{E}}, \texttt{\qt{E}}&&\texttt{n}, h, w&\textsc{left-arc}$_{12}$\\
\texttt{a}, \texttt{\qt{E}}&\texttt{\qt{E}}&\texttt{n}, h, w&\textsc{shift}$_{13}$\\
\texttt{\qt{E}}, \texttt{\qt{E}}, \texttt{n}&&E, h, w&\textsc{confirm}$_{14}$\\
\texttt{\qt{E}}, \texttt{\qt{E}}, \texttt{n}&&\texttt{p2}, h, w&\textsc{left-arc}$_{15}$\\
\texttt{a}, \texttt{\qt{E}}, \texttt{\qt{E}}&\texttt{n}, &\texttt{p2}, h, w&\textsc{no-arc}$_{16}$\\
\texttt{p}, \texttt{a}, \texttt{\qt{E}}&\texttt{\qt{E}}, \texttt{n}, &\texttt{p2}, h, w&\textsc{left-arc}$_{17}$\\
\texttt{p}, \texttt{a}&\texttt{\qt{E}}, \texttt{\qt{E}}, \texttt{n},&\texttt{p2}, h, w&\textsc{shift}$_{18}$\\

\texttt{\qt{E}}, \texttt{n}, \texttt{p2}&&h, w, s&\textsc{reduce}$\times 2_{19}$\\
\texttt{\qt{E}}, \texttt{n}, \texttt{p2}&&s, n, t&\textsc{confirm}$_{20}$\\
\texttt{\qt{E}}, \texttt{n}, \texttt{p2}&&\texttt{s}, n, t&\textsc{left-arc}$_{21}$\\
\texttt{\qt{E}}, \texttt{\qt{E}}, \texttt{n}&\texttt{p2}&\texttt{s}, n, t&\textsc{no-arc}$\times 3_{22}$\\
\texttt{p}, \texttt{a}&\texttt{\qt{E}}, \texttt{\qt{E}}, \texttt{n}&\texttt{s}, n, t&\textsc{left-arc}$\times 2_{23}$\\
&\texttt{p}, \texttt{a}, \texttt{\qt{E}}&\texttt{s}, n, t&\textsc{shift}$_{24}$\\
\texttt{n}, \texttt{p2}, \texttt{s}&&n, t, s2&\textsc{confirm}$_{25}$\\
\texttt{n}, \texttt{p2}, \texttt{s}&&\texttt{-}, t, s2&\textsc{shift}$_{26}$\\
\texttt{p2}, \texttt{s}, \texttt{-}&&t, s2, a2&\textsc{reduce}$_{27}$\\
\texttt{p2}, \texttt{s}, \texttt{-}&&s2, a2, p3&\textsc{confirm}$_{28}$\\
\texttt{p2}, \texttt{s}, \texttt{-}&&\texttt{s2}, a2, p3&\textsc{left-arc}$_{29}$\\
\texttt{n}, \texttt{p2}, \texttt{s}&\texttt{-}&\texttt{s2}, a2, p3&\textsc{no-arc} $\times 5_{30}$\\
\texttt{p}&\texttt{a}, \texttt{\qt{E}}, \texttt{\qt{E}}&\texttt{s2}, a2, p3&\textsc{left-arc}$_{31}$\\
&\texttt{p}, \texttt{a}, \texttt{\qt{E}}&\texttt{s2}, a2, p3&\textsc{shift}$_{32}$\\
\texttt{s}, \texttt{-}, \texttt{s2}&&a2, p3&\textsc{confirm}$_{33}$\\
\texttt{s}, \texttt{-}, \texttt{s2}&&\texttt{a2}, p3&\textsc{shift}$_{34}$\\
\texttt{-}, \texttt{s2}, \texttt{a2}&&p3&\textsc{confirm}$_{35}$\\
\texttt{-}, \texttt{s2}, \texttt{a2}&&\texttt{p3}&\textsc{left-arc}$_{36}$\\
\texttt{s}, \texttt{-}, \texttt{s2}&\texttt{a2}&\texttt{p3}&\textsc{right-arc}$_{37}$\\
\texttt{s2}, \texttt{a2}, \texttt{p3}&&&\textsc{shift}$_{38}$\\
\end{tabular}}

\caption{\label{table-gold-transitions-examples} Sequence of gold transitions to obtain the \textsc{amr} graph for the sentence \qt{When the prince arrived on the Earth, he was surprised not to see any people}, introduced in Figure \ref{f-amr-example}. For brevity, we represent words (and concepts) by their first character (plus an index if it is duplicated) and we only show the top three words for $\lambda_1$, $\lambda_2$ and $\beta$.
Steps from 20 to 23(2) and from 28 to 31 manage the reentrant edges for  \texttt{prince} (\texttt{p}) from \texttt{surprise-01} (\texttt{s}) and \texttt{see-01} (\texttt{s2}).
}
\end{table}

\subsection{Training the classifiers}

The algorithm relies on three classifiers: (1) a transition classifier, $T_c$, that learns the set of transitions introduced in \S \ref{section-formalization}, (2) a relation classifier, $R_c$, to predict the label(s) of an edge when the selected action is a \textsc{left-arc}, \textsc{right-arc} or \textsc{multiple-arc} and (3) a hybrid process (a concept classifier, $C_c$, and a rule-based system) that determines which concept to create when the selected action is a \textsc{confirm} or \textsc{breakdown}. 

\paragraph{Preprocessing} Sentences are tokenized and aligned with the concepts using J\textsc{amr} \cite{flanigan2014discriminative}. For lemmatization, tagging and dependency parsing we used UDpipe \cite{straka2016udpipe} and its English pre-trained model \cite{zeman2017conll}. Named Entity Recognition is handled by Stanford CoreNLP \cite{manning-EtAl:2014:P14-5}.

\paragraph{Architecture} We use feed-forward neural networks to train the tree classifiers. The transition classifier uses 2 hidden layers (400 and 200 input neurons) and the relation and concept classifiers use 1 hidden layer (200 neurons). The activation function in hidden layers is a $relu(x)$=$max(0,x)$ and their output is computed as $relu(W_i \cdot \vec{x}_i + b_i)$ 
where $W_i$ and $b_i$ are the weights and bias tensors to be learned and $\vec{x}_i$ the input at the $i$th hidden layer. The output layer uses a $\mathit{softmax}$ function, computed as
$P(y=s|\vec{x}) = \frac{e^{\vec{x^T}\theta_s}} {\sum_{s'=1}^{S} e^{\vec{x^T}\theta_{s'}}}$. All classifiers are trained in mini-batches (size=32), using Adam \cite{kingma2014adam} (learning rate set to $3e^{-4}$), early stopping (no patience) and dropout \cite{srivastava2014dropout} (40\%). The classifiers are fed with features extracted from the preprocessed texts. Depending on the classifier, we are using different features. These are summarized in Appendix \ref{appendix-suplemental-material} (Table \ref{table-set-of-features}), which also describes (\ref{appendix-design-decisions}) other design decisions that are not shown here due to space reasons.

\subsection{Running the system}

At each parsing configuration, we first try to find  a multiword concept or entity that matches the head elements of $\beta$, to reduce the number of \textsc{breakdown}s, which turned out to be a difficult transition to learn (see \S \ref{section-results}). This is done by looking at a lookup table of multiword concepts\footnote{The most frequent subgraph.} seen in the training set and a set of rules, as introduced in \cite{damonte-cohen-satta:2017:EACLlong,gildea-satta-cl17}.

We then invoke $T_c$ and call the corresponding subprocess when an additional concept or edge-label identification task is needed.

\paragraph{Concept identification} If the word at the top of $\beta$ occurred more than 4 times in the training set, we call a supervised classifier to predict the concept. Otherwise, we first look for a word-to-concept mapping in a lookup table. If not found, if it is a verb, we generate the concept \texttt{lemma-01}, and otherwise \texttt{lemma}.

\paragraph{Edge label identification} The classifier is invoked every time an edge is created. We use the list of valid \texttt{ARG}s allowed in propbank framesets by \newcite{damonte-cohen-satta:2017:EACLlong}. Also, if \texttt{p} and \texttt{o} are a propbank and a non-propbank concept, we restore inverted edges of the form \texttt{o} $\xrightarrow{\texttt{l-of}}$ \texttt{p} as \texttt{o} $\xrightarrow{\texttt{l}}$ \texttt{p}.

\section{Methods and Experiments}

\paragraph{Corpus} We use the LDC2015E86 corpus and its official splits: 16\,833 graphs for training, 1\,368 for development and 1\,371 for testing. The final model is only trained on the training split.

\paragraph{Metrics} We use Smatch \cite{cai2013smatch} and the metrics from \newcite{damonte-cohen-satta:2017:EACLlong}.\footnote{It is worth noting that the calculation of Smatch and metrics derived from it suffers from a random component, as they involve finding an alignment between  predicted and gold graphs with an approximate algorithm that can produce a suboptimal solution. Thus, as in previous work, reported Smatch scores may slightly underestimate the real score.}

\paragraph{Sources} The code and the pretrained model used in this paper can be found at \url{https://github.com/aghie/tb-amr}.

\subsection{Results and discussion}\label{section-results}

Table \ref{f-dev-T} shows accuracy of $T_c$ on the development set. \textsc{confirm} and \textsc{reduce} are the easiest transitions, as local information such as POS-tags and words are discriminative to distinguish between content and function words. \textsc{breakdown} is the hardest action.\footnote{This transition was trained/evaluated for non named-entity words that generated multiple nodes, e.g. father, that maps to \texttt{have-rel-role-91} \texttt{:ARG2} \texttt{father}.} In early stages of this work, we observed that this transition could learn to correctly generate multiple-term concepts for named-entities that are not sparse (e.g. countries or people), but failed with sparse entities (e.g. dates or percent quantities). Low performance on identifying them negatively affects the edge metrics, which require both concepts of an edge to be correct. Because of this and to identify them properly, we use the mentioned complementary rules to handle named entities. 
\textsc{right-arc}s are harder than \textsc{left-arc}s, although the reason for this issue remains as an open question for us. The performance for \textsc{no-arc}s is high, but it would be interesting to achieve a higher recall at a cost of a lower precision, as predicting \textsc{no-arc}s makes the transition sequence longer, but could help identify more distant reentrancy. The accuracy of $T_c$ is $\sim$86\%. The accuracy of $R_c$ is $\sim$79\%. We do not show the detailed results since the number of classes is too high. $C_c$ was trained on concepts occurring more than 1 time in the training set, obtaining an accuracy of $\sim$83\%. The accuracy on the development set with all concepts was $\sim$77\%.

\begin{table}[t]
\centering

\begin{tabular}{lccc}
 \bf Action &\bf Prec.&\bf Rec.&\bf F-score\\ 
\textsc{left-arc} &81.62&87.73&84.57\\
\textsc{right-arc} &75.53&78.71&77.08\\
\textsc{multiple-arc} &00.00&00.00&00.00\\
\textsc{shift} &80.44&81.11&80.77\\
\textsc{no-arc} &89.71&86.71&88.18\\
\textsc{confirm} &84.91&96.11&90.16\\
\textsc{reduce} &96.77&91.53&94.08\\
\textsc{breakdown} &85.09&50.23&63.17\\
\end{tabular}
\caption{\label{f-dev-T} $T_c$ scores on the development set.}
\end{table}

Table \ref{f-test-SOTA} compares the performance of our systems with state-of-the-art models on the test set. \textsc{amr-covington} obtains state-of-the-art results for all the standard metrics. It outperforms the rest of the models when handling reentrant edges. It is worth noting that D requires an additional classifier to handle a restricted set of reentrancy and P uses up to five classifiers to build the graph.

\begin{table}[t]
\centering

\begin{tabular}{lccc|ccc}
 \bf Metric & \bf F& \bf W & \bf F' &\bf D& \bf P &\bf Ours \\ 
Smatch &58&63&67&64&64&64\\

Unlabeled &61&69&69&69&-&68\\

No-WSD &58&64&68&65&-&65\\

NER &75&75&79&83&-&83\\

Wiki&0&0&75&64&-&70\\
Negations &16&18&45&48&-&47\\ 
Concepts &79&80&83&83&-&83\\
Reentrancy &38&41&42&41&-&44\\
SRL &55&60&60&56&-&57\\
\end{tabular}
\caption{\label{f-test-SOTA} F-score comparison with F~\cite{flanigan2014discriminative}, W~\cite{wang2015transition}, F'~\cite{flanigan2016cmu}, D~\cite{damonte-cohen-satta:2017:EACLlong}, P~\cite{peng-aaai18}. 
D, P and our system are left-to-right transition-based.
}
\end{table}

\paragraph{Discussion} In contrast to related work that relies on \emph{ad-hoc} procedures, the proposed algorithm handles cycles and reentrant edges natively. This is done by just removing the original constraints of the arc transitions in the original \newcite{covington2001fundamental} algorithm. The main drawback of the algorithm is its computational complexity. The transition system is expected to run in $\mathcal{O}(n^2)$, as the original Covington parser. There are also collateral issues that impact the real speed of the system, such as predicting the concepts in a supervised way, given the large number of output classes (discarding the less frequent concepts the classifier needs to discriminate among more than 7\,000 concepts). In line with previous discussions \cite{damonte-cohen-satta:2017:EACLlong}, it seems that using a supervised feed-forward network to predict the concepts does not lead to a better overall concept identification with respect of the use of simple lookup tables that pick up the most common node/subgraph.
Currently, every node is kept in $\lambda$, and it is available to be part of new edges. We wonder if only storing in $\lambda$ the head node for words that generate multiple-node subgraphs (e.g. for the word father that maps to \texttt{have-rel-role-91} \texttt{:ARG2} \texttt{father}, keeping in $\lambda$ only the concept \texttt{have-rel-role-91}) could be beneficial for \textsc{amr-covington}.

As a side note, current \textsc{amr} evaluation involves elements such as neural network initialization, hooks and the (sub-optimal) alignments of evaluation metrics (e.g. Smatch) that introduce random effects that were difficult to quantify for us.

\section{Conclusion}

We introduce \textsc{amr-covington}, a non-projective transition-based parser for unrestricted \textsc{amr}. The set of transitions handles reentrancy natively. Experiments on the LDC2015E86 corpus show that our approach obtains results close to the state of the art and a good behavior on re-entrant edges.

As future work, \textsc{amr-covington} produces sequences of \textsc{no-arc}s which could be shortened by using non-local transitions
\cite{qi-manning:2017:Short,2017arXiv171009340F}. Sequential models have shown that fewer hooks and lookup tables are needed to deal with the high sparsity of \textsc{amr} \cite{ballesteros-alonaizan:2017:EMNLP2017}. Similarly, \textsc{bist-covington} \cite{vilares2017non} could be adapted for this task.

\section*{Acknowledgments}

This work is funded from the European
Research Council (ERC), under the European
Union's Horizon 2020 research and innovation
programme (FASTPARSE, grant agreement No
714150), from the TELEPARES-UDC project
(FFI2014-51978-C2-2-R) and the ANSWER-ASAP project (TIN2017-85160-C2-1-R) from MINECO, and from Xunta de Galicia (ED431B 2017/01). 
We gratefully acknowledge NVIDIA Corporation for the donation of a GTX Titan X GPU.

\bibliographystyle{acl_natbib}
\bibliography{biblio}

\appendix

\section{Supplemental Material}\label{appendix-suplemental-material}

Table \ref{table-set-of-features} indicates the features used to train the different classifiers. Concept features are randomized  by $\epsilon=2e^{-3}$ to an special index that refers to an \texttt{unknown concept} during the training phase. This helps learn a generic embedding for unseen concepts in the test phase. Also, concepts that occur only one time in the training set are not considered as output classes by $C_c$.

\begin{table}[hbtp]
\centering
\small{
\begin{tabular}{lccc}
 \bf Features & \bf $T_c$& \bf $R_c$& \bf $C_c$\\ 
\multicolumn{4}{l}{\emph{From} $\beta_0, \beta_1, \lambda_{1_0}, \lambda_{1_1}$}\\
\textsc{pos}&\checkmark&\checkmark&\checkmark\\
\textsc{w}&\checkmark&\checkmark&\checkmark\\
\textsc{ew} &\checkmark&\checkmark&\checkmark\\
\textsc{c}&\checkmark&\checkmark&\checkmark\\
\textsc{entity}&\checkmark&\checkmark&\checkmark\\

\textsc{lm}$_h$&\checkmark&\checkmark&\\
\textsc{lm}$_c$&\checkmark&\checkmark&\\
\textsc{lm}$_{cc}$&\checkmark&\checkmark&\\
\textsc{rm}$_h$&\checkmark&\checkmark&\\
\textsc{rm}$_c$&\checkmark&\checkmark&\\
\textsc{rm}$_{cc}$&\checkmark&\checkmark&\\
\textsc{nh}, \textsc{nc} &\checkmark&\checkmark&\\ 
\textsc{depth} &\checkmark&\checkmark&\checkmark\\
\textsc{npunkt}&\checkmark&\checkmark&\\
\textsc{hl}&\checkmark&\checkmark&\\
\textsc{ct}&\checkmark&\checkmark&\\
\textsc{g}&&&\checkmark\\
\multicolumn{4}{l}{\emph{Labels from the predicted}}\\
\multicolumn{4}{l}{\emph{dependency tree}}\\
\textsc{d}$(b_j,i_k)\ \forall j,k\ j \in [0,1]$&\checkmark&\checkmark&\checkmark\\
$\wedge\ k \in[0,1,2,3]$\\
\end{tabular}}
\caption{\label{table-set-of-features} Set of proposed features 
for each classifier. \textsc{pos}, \textsc{w}, \textsc{c}, \textsc{entity} are  part-of-speech tag, word, concept and entity embeddings. \textsc{ew} are pre-trained external word embeddings, fine-tuned during the training phase (http://nlp.stanford.edu/data/glove.6B.zip, 100 dimensions). \textsc{lm} and \textsc{rm} are the leftmost and the rightmost function; and h, c, cc represent head, child and grand-child concepts of a concept; so, for example, \textsc{lm}$_c$ stands for the leftmost child of the concept. \textsc{nh} and \textsc{nc} are the number of heads and children of a concept. \textsc{npunkt} indicates the number of \qt{.}, \qt{;}, \qt{:}, \qt{?}, \qt{!} that have already been processed. \textsc{hl} denotes the labels of the last assigned head. \textsc{ct} indicates the type of concept (constant, propbank frameset, other). 
\textsc{g}  indicates if a concept was generated using a \textsc{confirm} or \textsc{breakdown}.
\textsc{d} denotes the dependency label existing in the dependency tree between the word at the $j$th position in $\beta$ and the $k$th last in $\lambda_1$ and vice versa. The word that generated a concept is still accessible after creating the concept.}
\end{table}

Internal and external (from GloVe) word embedding sizes are set to 100. The size of the concept embedding is set to 50. The rest of the embeddings size are set to 20. The weights are initialized to zero.

\section{Additional design decisions}\label{appendix-design-decisions}

We try to list more in detail the main hooks and design decisions followed in this work to mitigate the high sparsity in Abstract Meaning Representation which, at least in our experience, was a struggling issue. These decisions mainly affect the mapping from words to multiple-concept subgraphs.

\begin{itemize}[leftmargin=*]
\item We identify named entities and nationalities, and update the training configurations to generate the corresponding subgraph by applying a set of hooks.\footnote{The hooks are based on the text file resources for countries, cities, etc, released by \citet{damonte-cohen-satta:2017:EACLlong} and an analysis of how named entities are generated in the training/development set.} The intermediate training configurations are not fed as samples to the classifier. On early experiments it was observed that the \textsc{breakdown} transition could acceptably learn non-sparse named entities (e.g. countries and nationalities), but failed on the sparse ones (e.g. dates or money amounts). By processing the named entities with hooks instead, the aim was to make the parser familiar with the parsing configurations that are obtained after applying the hooks.

\item Additionally, named-entity subgraphs and subgraphs coming from phrases (involving two or more terms) from the training set are saved into a lookup table. The latter ones had little impact.

\item We store in a lookup table some single-word expressions that generated multiple-concept subgraphs in the training set, based on simple heuristics. We store words that denote a negative expression (e.g. undecided that maps to \texttt{decide-01} \texttt{:polarity} \texttt{-}). We store words that always generated the same subgraph and occurred more than 5 times. We also store capitalized single words that were not previously identified as named entities.

\item We use the verbalization list from \newcite{wang2015transition} (another lookup table).

\item When predicting a \textsc{confirm} or \textsc{breakdown} for an uncommon word, we check if that word was mapped to a concept in the training set. If not, we generate the concept \texttt{lemma-01} if it is a verb, otherwise \texttt{lemma}.

\item Dates formatted as YYMMDD or YYYYMMDD are identified using a simple criterion (sequence of 6 or 8 digits) and transformed into YYYY-MM-DD on the test phase, as they were consistently misclassified as integer numbers in the development set.

\item We apply a set of hooks similar to \cite{damonte-cohen-satta:2017:EACLlong} to determine if the predicted label is valid for that edge.

\item We forbid to generate the same concept twice consecutively. Also, we set $\alpha=4$ for \textsc{breakdown}$_\alpha$.

\item If a node is created, but it is not attached to any head node, we post-process it and connect to the root node.

\item We assume multi-sentence graphs should contain sentence punctuation symbols. If we predict a multi-sentence graph, but there is no punctuation symbol that splits sentences, we post-process the graph and transform the root node into an \textsc{and} node.

\end{itemize}

\end{document}